# How do walkers avoid a mobile robot crossing their way?

Christian Vassallo [a], Anne-Hélène Olivier [b,c], Philippe Souères [a], Armel Crétual [b], Olivier Stasse [a] and Julien Pettré [c]

a) CNRS, LAAS, 7 Avenue du colonel Roche, F-31400 Toulouse, France Univ de Toulouse, LAAS, F-31400 Toulouse, France
b) M2S lab (Mouvement Sport Santé), University Rennes 2 — ENS Rennes — UEB, avenue Robert Schuman, Campus de Ker Lann, 35170 Bruz, France
c) MimeTIC team, INRIA Rennes, Centre de Rennes Bretagne Atlantique, Campus universitaire de Beaulieu, 35042 Rennes, France

**E-mail addresses**
cvassall@laas.fr (C. Vassallo),
anne-helene.olivier@univ-rennes2.fr (A.H. Olivier),
soueres@laas.fr, (P. Souères),
armel.cretual@univ-rennes2.fr (A. Crétual),
ostasse@laas.fr (O. Stasse),
julien.pettre@inria.fr (J. Pettré)

# ABSTRACT

Robots and Humans have to share the same environment more and more often. In the aim of steering robots in a safe and convenient manner among humans it is required to understand how humans interact with them. This work focuses on collision avoidance between a human and a robot during locomotion. Having in mind previous results on human obstacle avoidance, as well as the description of the main principles which guide collision avoidance strategies, we observe how humans adapt a goal-directed locomotion task when they have to interfere with a mobile robot. Our results show differences in the strategy set by humans to avoid a robot in comparison with avoiding another human. Humans prefer to give the way to the robot even when they are likely to pass first at the beginning of the interaction.

# INTRODUCTION

Robots and humans will have to share the same environment in a near future [6, 11]. To this end, roboticists must guarantee safe interactions between robots and humans during locomotion tasks. In this direction, the following paper studies how humans behave to avoid a mobile robot crossing their way.

There is an extensive literature describing how walkers avoid collisions. Several studies considered how walkers step over [16] or circumvent [18] static obstacles. More recent ones focused on how humans avoid each other. It was shown that walkers are able to predict the risk of collision since they adapt their motion only if the future crossing distance is below a certain threshold [13]. This future distance is increased before the crossing point and maintained constant during a regulation phase, demonstrating anticipation in avoidance [13]. Trajectory adaptations are performed both in speed and orientation [8, 14]: they depend on the crossing angle and the walking speed [7]. These strategies do not maximize smoothness [1], they result from a compromise between safety and energy [8]. Moreover, these adaptations depend more on situations than personal characteristics [10].

The crossing order during collision avoidance is an interesting parameter to consider. Indeed, it has been shown that trajectory adaptations are collaboratively performed [13] but are role-dependent. The walker giving way (2nd at the crossing) contributes more than the one passing first. This role attribution appears to contribute positively before the interaction [10, 14] and can be predicted with 95% confidence at 2.5m before crossing, even before any adaptation [10].

Studies resulted into simulation models of navigation and interaction. Warren and Fajen [20] proposed to model the walker and the environment as coupled dynamical systems: the walker paths result from all the forces acting on them, where goals are considered as attractors and obstacles as repellors. This model is based on the distance to the goal and to the obstacles as well as the sign of change of the bearing angle. An integration of the bearing angle theory into some artificial vision system for crowd simulation was proposed by Ondrej et al. [15].

These studies reached common conclusions about the human ability to accurately estimate the situation (crossing order, risk of collision, adaptations), and considered interactions with a moving object. The kinematics of adaptation by a walker avoiding a moving obstacle (a mannequin mounted on a rail) are studied in [3, 5]. Trajectories crossing at 45° resulted into adaptations both in the antero-posterior and medio-lateral planes, with successive anticipation and clearance phases [5]. Analysis is based on the notion of personal space modeled as a free elliptic area around walkers. When trajectories are collinear (the mannequin comes from front), a 2-step avoidance strategy is observed: participants first adapt their locomotion in heading, followed by speed [3]. These experiments with mannequins were not designed to study the question of the crossing order (either participants were forced to give way, or there is order in a collinear situation). Other studies investigated human interactions with robots. It was shown that it is easier to understand and predict the behavior of robots if they are human-like [2, 12]. Some studies demonstrated that human-like behaviors [4, 9] improve on many levels the performance of human-robot collaboration. Nevertheless, the benefit of programming a robot with human-like capabilities to move and avoid collision with a human walker has not been demonstrated yet.

In this paper, we use a robot to interfere with a pedestrian. We control the robot to reproduce similar kinematic conditions of interaction than the ones studied in [13] (in terms of relative angle, position, and velocity) and apply a similar analysis. While the nature of the interaction is changed, we show differences in the strategies set by participants with respect to previous observations.

# MATERIALS AND METHODS

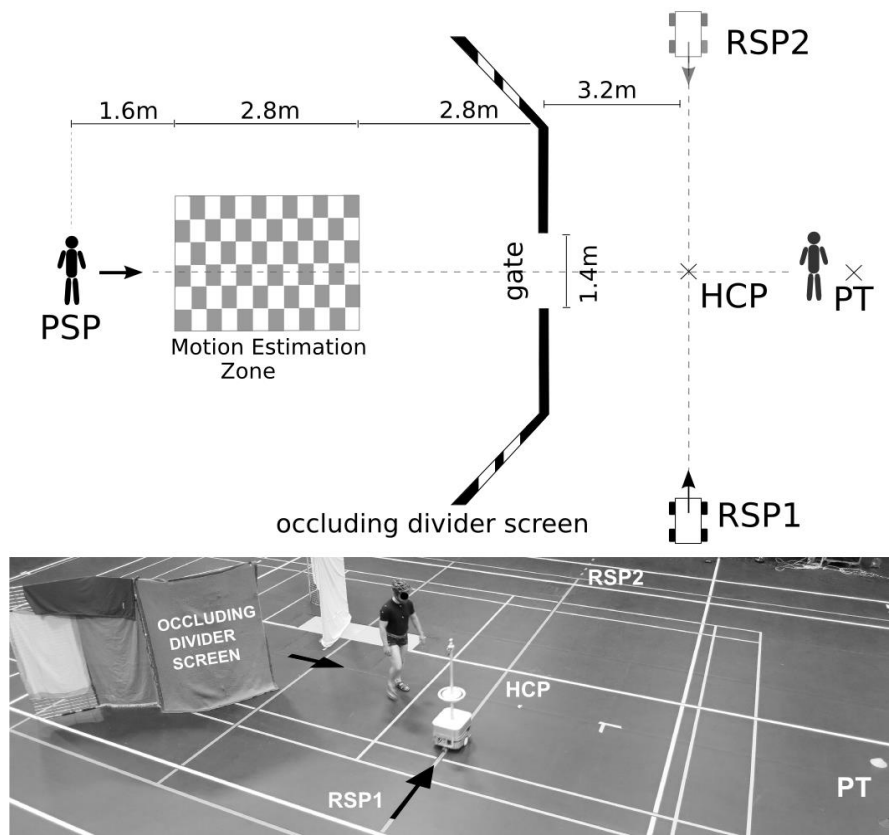

*Figure 1: Experimental apparatus and task. In this trial the robot was moving from RSP1 to RSP2. Participant decided to pass behind the robot.*

## Participants

Seven volunteers participated in the experiment (1 woman and 6 men). They were 26.1 (±5.4) years old and 1.78m tall (±0.21). They had no known vestibular, neurological or muscular pathology that would affect their locomotion. All of them had normal or corrected sight and hearing. Participants gave written and informed consent before their inclusion in the study. The experiments respect the standards of the Declaration of Helsinki (rev. 2013), with formal approval of the ethics evaluation committee Comité d'Evaluation Ethique de l'Inserm (IRB00003888, Opinion number 13-124) of the Institut National de la Santé et de la Recherche Médicale, INSERM, Paris, France (IORG0003254, FWA00005831).

## Apparatus

The experiment took place in 40m x 25m gymnasium. The room was separated in two areas by 2m high occluding walls forming a gate in the middle (Figure 1). Four specific positions were identified: the participant starting position PSP, the participant target PT, and two robot starting positions RSP1 and RSP2. A specific zone between PSP and the gate is named Motion Estimation Zone MEZ. MEZ is far enough from PSP for the participants to reach their comfort velocity before entering the MEZ. The point of intersection between the robot path [RSP1, RSP2] and the participant path [PSP, PT] is named Hypothetical Crossing Point HCP. It is computed by hypothesizing that there is no adaptation of the participant trajectory.

### Participant Task

Participants were asked to walk at their preferred speed from PSP to PT by passing through the gate. They were told that an obstacle is moving over the gate and could interfere with them. One experimental trial corresponds to one travel from PSP to PT.

### Recorded data

3D kinematic data were recorded using the motion capture Vicon-MX system (120Hz). Reconstruction was performed using Vicon-Blade and computations using Matlab (Mathworks ®). The experimental area was covered by 15 infrared cameras. The global position of participants was estimated as the middle point of reflective markers set on the shoulders (acromion anatomical landmark). The stepping oscillations were filtered out by applying a Butterworth low-pass filter (2nd order, dual pass, 0.5Hz cut-off frequency).

### Robot Behavior

We used *RobuLAB10* robot from the *Robosoft* company (dimension: 0.45 x 0.40 x 1.42m, weight 25 Kg, maximal speed ~3 $m.s^{-1}$). The robot position was detected as the center point in its base. We programmed the robot to execute a straight trajectory between RSP1 and RSP2 at constant speed (1.4 $m.s^{-1}$). The robot was controlled to generate specific interactions with the participant. In particular, the robot was either: a) on a full collision course (reach HCP at the same time than the participant), b) on a partial collision course (the robot reaches HCP slightly before or after the participant), or c) not on a collision course. To this end, we measured the participant's speed through MEZ and estimated the time *t_hcp* when HCP was reached. We deduced the time *t_rs* at which the robot should start to reach HCP at *t_hcp*. We finally added an offset Delta to *t_rs*, randomly selected from the range [-0.6, 0.6s], to create the desired range of interactions.

### Experimental plan

Each participant performed 40 trials. Robot starting position (50% in RSP1, 50% in RSP2) was randomized among the trials. To introduce a bit of variability, in 4 trials the robot did not move and the participant did not have to react. Only the 36 trials with potential interaction were analyzed.

# ANALYSIS

### Kinematic data

For each trial we computed *t_rob*, the time at which the robot reaches its constant cruise speed (when the acceleration amplitude was below a fixed threshold, 0.003 $m.s^{-2}$), and *t_cross*, the time of closest approach between the participant and the robot.

The crossing configuration and the risk of future collision was estimated by the Minimal Predicted Distance, noted mpd [13]. mpd gives, at each time step, the future distance of closest approach if both the robot and the participant keep a constant speed and direction. Thus, a variation of mpd means that the participant is performing adaptations.

We introduced a signed definition of the mpd, noted *smpd*. The sign of this function depends on who, among the participant and the robot, is likely to reach HCP first (still assuming constant motion).

smpd is positive if the participant should arrive as first and is negative otherwise. A change of sign of smpd means that the future crossing order between the robot and the participant is switched. Our study focuses on the section of data when adaptations are performed: smpd is normalized in time by resampling the function at 100 intervals between *t_rob* (time 0%) and *t_cross* (time 100%). Quantity of adaptation is computed as the absolute value of the difference between smpd(*t_rob*) and smpd(*t_cross*).

## Statistics

Statistics were performed using Statistica (Statsoft ®). All effects were reported at p<0.05. Normality was assessed using a Kolmogorov-Smirnov test. Depending on the normality, values are expressed as median (M) or mean ±SD. Wilcoxon signed-rank tests were used to determine differences between values of smpd at *t_rob* and *t_cross*. The influence of the crossing order evolution on smpd values was assessed using a Kruskal-Wallis test with post hoc Mann-Whitney tests for which a Bonferroni correction was applied: all effects are reported at a 0.016 level of significance (0.05/3). Finally, we used a Mann-Whitney test to compare the crossing distance depending on the final crossing order.

# RESULTS

We considered 243 trials. Nine of the 252 were removed from analysis as the robot failed to start. Figure 2 depicts the evolution of smpd for all trials.

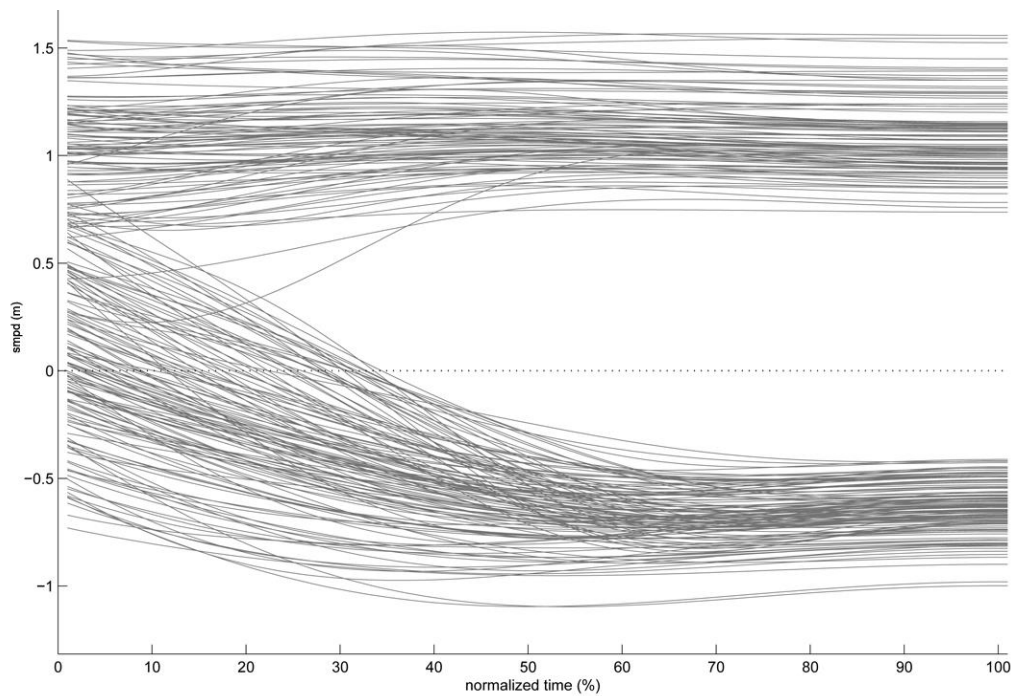

*Figure 2: smpd plots for all the 243 trials, after resampling over the interaction period of time [t_rob, t_cross]*

The sign of smpd at *t_cross* shows that the participants gave way in 58% of trails, otherwise they passed first. Combining this information with the sign of smpd at *t_rob* (beginning of interaction), we could evaluate if participants switched their role during interaction, i.e., change the crossing order. We divided trials into four categories depending on the sign of smpd at *t_rob* and at *t_cross*: PosPos, NegNeg, PosNeg and NegPos. For example, the PosNeg category is for trials with smpd(*t_rob*)>0 and smpd(*t_cross*)<0.

All the trials are distributed among those categories in the following way: PosPos=104 trials (43%), NegNeg=69 trials (28%), PosNeg=70 trials (29%), NegPos=0 trials (0%). Examples of corresponding trajectories for each category are illustrated in Figure 3. In 29% of cases, participants were likely to pass first at the crossing point but adapted their trajectory to finally give way to the robot. However, the opposite case in which participants would be likely to give way and finally would pass first was never observed.

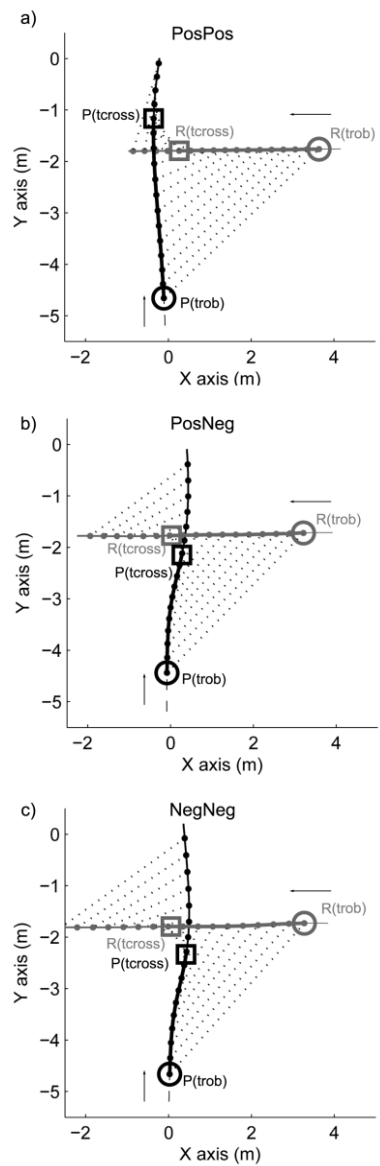

*Figure 3. 3 examples of participant-robot trajectories, for PosPos (top), PosNeg (middle) and NegNeg (bottom) category of trial. The part of the trajectory corresponding to the interaction [t_rob, t_cross] is bold. Time equivalent participant-robot positions are linked by dotted line.*

The mean evolution of smpd in each category is shown in Figure 4a and its time derivative in Figure 4b. From these curves we can distinguish the reaction period from the regulation one as defined by [13]. They respectively correspond to periods during which participants perform adaptations (smpd variates) or consider collision to be avoided (the derivative is null, and may even variate in the opposite direction).

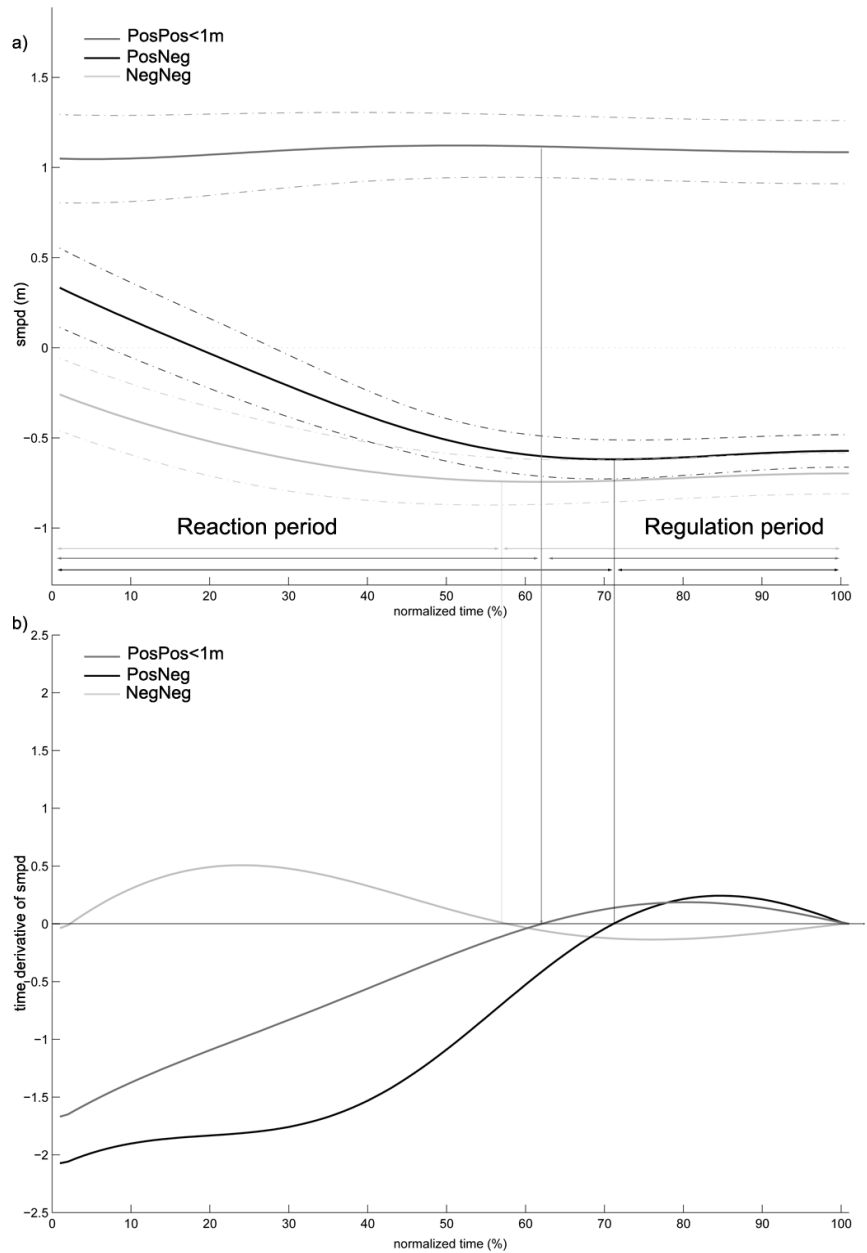

*Figure 4: (a) Mean evolution of smpd for each category of trial ±1 SD. (b) Time derivative of the mean smpd.*

Figure 5 shows comparisons between smpd($t\_rob$) and smpd($t\_cross$), as well as distances of closest approach with respect to each of the three non-empty categories, and the crossing distance depending on the crossing order.

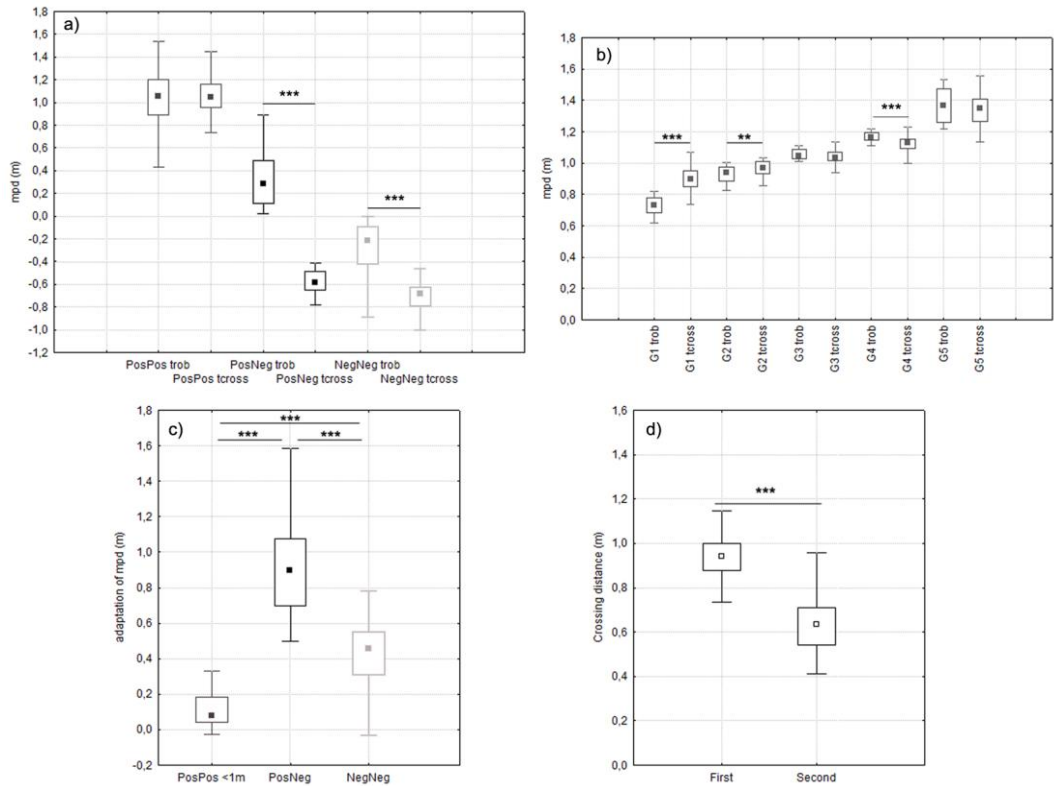

*Figure 5: a) Comparison between initial and final values of smpd at t_rob and t_cross for all trials of each non-empty category (PosPos, PosNeg, NegNeg). A significant difference in values means that adaptations were made to the trajectory by the participant (\*\*p<0.01, \*\*\*p<0.001). b) Comparison between initial and final values of smpd at t_rob and t_cross for all the trials of each subgroup of the PosPos category. c) Total variation for smpd over the interaction. d) Minimum distance observed between the robot and the participant at t_cross, with trial grouped by passing order of the participant.*

**PosPos trials.** Difference between smpd at *t_rob* (1.05m) and at *t_cross* (1.04m) is at the limit of significance (p=0.052), cf Figure (5-a). Participants performed no or few adaptations to their trajectory. We detail this result by ordering PosPos trials by increasing the smpd(*t_rob*) value and dividing them into 5 subgroups of the same size (cf Figure 5-b). Results show that there was a significant increase of smpd between *t_rob* and *t_cross* for subgroups 1 and 2 (respectively Msmpdtrob=0.73m, Msmpdtcross=0.89m, Z=3.92, p<0.0001, r=0.88 and Msmpdtrob=0.94m, Msmpdtcross=0.97m, Z=3.17, p<0.01, r=0.71). For subgroups 3 and 5, there was no significant smpd variation between *t_rob* and *t_cross* (respectively Msmpdtrob=1.05m, Msmpdtcross=1.04m, p=0.14 and Msmpdtrob=1.37m, Msmpdtcross=1.35m, p=0.11). For subgroup 4 there was a significant decrease of smpd between *t_rob* and *t_cross* (Msmpdtrob=1.18m, Msmpdtcross=1.16m, Z=2.42, p<0.05, r=0.54). This result suggests that when the initial value of smpd(*t_rob*) is lower than 1m, participants significantly adapted their trajectory.

In the sequel we neglected subgroups 3, 4 and 5 since they do not capture avoidance adaptations. The updated Pos-Pos group is then named PosPos<1m. Its median values of smpd(*t_rob*) and smpd(*t_cross*) were respectively 0.81m and 0.94m.

**PosNeg trials.** Results showed a significant difference of smpd between *t_rob* and *t_cross* (Msmpdtrob=0.29m, Msmpdtcross=-0.58m, Z=7.23, p<0.0001, r=0.86). When participants initially were likely to pass first with an existing risk of collision, they finally decided to give way.

**NegNeg trials.** Results showed a significant difference of smpd between *t_rob* and *t_cross* (Msmpdtrob=-0.22m, Msmpdtcross=-0.68m, Z=7.21, p<0.0001, r=0.87). When participants were likely to give way, still with an existing risk of collision, they increased the future distance of closest approach while maintaining the crossing order.

**Quantity of adaptation.** The more smpd varies, the more adaptation is performed (Figure 5c). There is an influence of the group on the quantity of adaptation: H(2,180)=132.17, p<0.0001). Post hoc test showed that more adaptation is performed for trials of the PosNeg group (M=0.90m) than for NegNeg (M=0.46m). Both show more adaptation than for PosPos<1m (M=0.07mn), p<0.001.

**Crossing distance.** The average distance of closest approach was 0.83m (±0.27m). The crossing order influenced this distance (Figure 5d). It was higher when participants passed first (M=0.94m) (U=175, Z=-9.03, p<0.0001).

# DISCUSSION

We studied how humans adapt their trajectory to avoid collision with a moving robot. We considered situations which are similar to previous studies on human-human collision avoidance and we based our analysis was based on the concept of mpd introduced in [13] to enable comparisons.

Results show that, concerning some aspects, humans have similar behavior when avoiding a robot or another human [13]. They accurately estimate the future risk of collision: they do not perform adaptations when smpd is initially high (smpd(*t_rob*)>1m). They also solve collision avoidance with anticipation. Figure 4 shows that smpd plateaus in the latter period of interaction, called the regulation phase. A constant smpd indicates that the avoidance maneuvers are over, and this happens significantly before *t_cross*.

A significant difference about crossing configuration concerns the crossing order inversions that was observed in 29% of the trials: participants were initially likely to pass first but adapted their trajectory to finally give way to the robot. This threshold can be defined so that smpd(*t_rob*) is around 0.8m, which is the value of smpd(*t_rob*) for the Pos-Pos<1m group: above this threshold, participants remain first, otherwise they change the crossing order to give way to the robot. There was an overlapping in the smpd(*t_rob*) values between Pos-Pos<1m and Pos-Neg groups. To confirm such a threshold, we compared these smpd(*t_rob*) values using a Mann-Whitney test. Results showed that smpd(*t_rob*) for Pos-Pos<1m trials (M=0.81m) was higher than for the Pos-Neg trials (M=0.30m) (U=104, Z=7.69, p<0.0001). We can link this 0.8m threshold to the crossing distance observed between two pedestrians in similar conditions [13] that was 0.81m.

The absence of inversion NegPos was also similarly observed and modeled by the bearing angle theory [11]. Assuming that the robot obeys the bearing angle model and the pedestrian contradicts it by re-inverting the sign of the bearing angle change, the two adaptations could cancel out each other, ending into failure.

Which factors may explain such an inversion of role? We can assume that changing role in the crossing order is somehow inefficient: no physiological or kinematic factor explains this observation. It is not relevant to explain this based on perceptual factors (can humans early perceive the role they are likely to have?). Indeed, inversion is observed in only one direction (Pos-Neg), and we observed that the risk of collision is correctly perceived. We interpret the role inversion as an extreme adaptation to preserve the personal space, directly related to the perceived risk of collision. Passing in front of the robot may be perceived as more dangerous. We suggest the following hypotheses to explain this feeling of danger. First, humans ignored how the robot was controlled and its goal. The ongoing situation cannot be easily predicted by pedestrians during short interactions. Our second hypothesis deals with the lack of experience of participants to interact with an autonomous system. Humans perform collision avoidance with other humans daily and they may expect collaboration from them. Human-Robot collisions occur indeed very rarely in real life, which can be linked to theories describing the effect of uncertainty in motor control [17, 19]. Our third hypothesis is about the danger of getting hurt: visually the robot looks quite heavy, metallic and compact. Colliding with it could certainly hurt legs which makes reasonable that participants prefer to adopt a safer behavior than when crossing a human.

## CONCLUSION

In this paper, we observed that human-robot collision avoidance has similarities with human-human interactions (estimation of collision risk, anticipation) but also leads to some major differences. Humans preferentially give way to the robot, even though this choice is not optimal to avoid collision. We interpret this behavior based on the notion of perceived danger and safety. This conservative strategy could be due to the lack of understanding of how the robot behaves and the lack of experience of such an interaction with an autonomous system. However, human always succeeded in avoiding the robot with anticipation and without aberrant reaction. We also raised questions about humans reactions facing a robot programmed to behave as a human. Would humans understand that the robot cooperates and adapt their own strategy accordingly?

The conclusion of this study opens doors for future research. A first direction is to better understand the possible effect of this notion of danger during human-robot interactions. What are the aspects that influence more human perception (velocity, shape, size)? A second direction is the design of safe robot motion amidst human walkers. How should the robot adapt to humans? Should it be collaborative with the risk of compensating human avoidance strategies? Should it be passive? We believe that robots should be first equipped with human abilities to early detect human avoidance strategy and adapt to it.

In the near future, we will continue our study of human-robot crossing interaction. We plan to equip the robot with collision avoidance capabilities that imitates human strategies, and investigate how participants adapt to this new attitude compared to a passive robot.